\documentclass[lettersize,journal]{IEEEtran}

\usepackage[utf8]{inputenc}
\usepackage[T1]{fontenc}    %
\usepackage{nicefrac}       %
\usepackage{microtype}      %
\usepackage{times}

\usepackage{multicol}
\usepackage[hidelinks]{hyperref}

\usepackage{standalone}
\IfStandalone{
\usepackage[top=2cm,bottom=2.5cm,left=2.5cm,right=2.5cm]{geometry}}{
}

\usepackage{mathtools}
\usepackage{amsthm,amsmath}
\usepackage{amssymb}

\usepackage{comment}
\usepackage{todonotes}
\presetkeys{todonotes}{inline}{}

\usepackage{tikz}
\usepackage{pgfplots}
\usepackage[ruled,vlined]{algorithm2e}
\DontPrintSemicolon

\usepackage[capitalize]{cleveref}
\usepackage[acronym]{glossaries}

\usepackage[font=footnotesize]{caption}
\usepackage[font=footnotesize]{subcaption}

\usepackage{xcolor}
\usepackage{multirow}
\usepackage{booktabs}
\usepackage{threeparttable}
\usepackage{makecell}
\usepackage{cleveref}
\Crefname{equation}{Eq.}{Eqs.}
\Crefname{figure}{Fig.}{Figs.}
\Crefname{tabular}{Tab.}{Tabs.}
\Crefname{definition}{Def.}{Defs.}
\Crefname{section}{Sec.}{Sects.}
\Crefname{theorem}{Thm.}{Thms.}

\usepackage{ifxetex}
\ifxetex
\usepackage[tracking=false,kerning=false,spacing=false]{microtype}
\usepackage[protrusion=true,tracking=false,kerning=false,spacing=false]{microtype}
\else
\fi
\usepackage{enumitem}

\newcommand{\agents}{\mathcal{A}}
\newcommand{\actionset}{U}
\newcommand{\action}{u}
\newcommand{\av}{\mathsf{av}}

\DeclareMathOperator*{\argmin}{arg\,min}

\newcommand{\bernoulli}{\mathsf{Bernoulli}}

\newcommand{\counterfactualrv}{\mathsf{C}}
\newcommand{\crealization}{\mathsf{c}}
\newcommand{\cfintensity}{\lambda} %

\newcommand{\coll}{\mathrm{coll}} %
\newcommand{\collseverity}{\mathsf{coll\_severity}}

\newacronym{cvar}{CVaR}{Conditional Value at Risk}
\newcommand{\dynamics}{f}

\newcommand{\episode}{e}
\newcommand{\episoderv}{\mathsf{E}}

\newcommand{\horizon}{T}

\newacronym{idm}{IDM}{Intelligent Driver Model}

\newacronym{mais}{MAIS}{Maximum Abbreviated Injury Scale}

\newcommand{\odd}{p}    %

\newacronym{poset}{poset}{partially ordered set}
\newcommand{\proofref}[1]{\ifthenelse{\boolean{submission}}{}{The proof can be found in \cref{#1}.}}

\newcommand{\policy}{\pi}

\newcommand{\param}{\theta}

\newcommand{\prob}{\mathbb{P}} %

\newcommand{\risk}{r}

\newcommand{\statespace}{X}
\newcommand{\state}{x}

\newcommand{\simplex}{\Delta}

\newcommand{\safetym}{\cfintensity^*} %
\newcommand{\highspeed}{\textsc{high speed}\xspace} %
\newcommand{\lowspeed}{\textsc{low speed}\xspace}

\newcommand{\tup}[1]{\langle#1\rangle}

\newacronym{av}{AV}{Autonomous Vehicle}
\newacronym{rss}{RSS}{Responsibility-Sensitive Safety}
\newacronym{odd}{ODD}{Operational Design Domain}
\newacronym{ne}{NE}{Nash Equilibrium}

\newcommand{\reals}{\mathbb{R}}

\newcommand{\bool}{\mathrm{Bool}}

\newtheorem{definition}{Definition}

\newtheorem{assumption}{Assumption}

\title{A Counterfactual Safety Margin Perspective on the Scoring of Autonomous Vehicles' Riskiness}
\author{Alessandro Zanardi$^{1}$,~\IEEEmembership{Graduate Student Member,~IEEE,} Andrea Censi$^{1}$,~\IEEEmembership{Member,~IEEE,} Margherita Atzei$^{2}$,\\Luigi Di Lillo$^{2,3}$~\IEEEmembership{Member,~IEEE,}, Emilio Frazzoli$^{1}$~\IEEEmembership{Fellow,~IEEE,}
\thanks{$^{1}$A. Zanardi, A. Censi, and E. Frazzoli are with the Institute for Dynamic Systems and Control, ETH Z\"urich, Switzerland\\\{azanardi,acensi,emilio.frazzoli\}@ethz.ch}
\thanks{$^{2}$M. Atzei and L. Di Lillo are with Property and Casualty Solutions, Reinsurance, Swiss Reinsurance Company, Ltd., Z\"urich, Switzerland.
\{margherita\_atzei,luigi\_dilillo\}@swissre.com}
\thanks{$^{3}$L. Di Lillo is a research affiliate with the Autonomous Systems Laboratory at Stanford University and a research collaborator of the Frazzoli Group at ETH Zurich.}
\thanks{This work was supported by the Swiss National Science Foundation under NCCR Automation, grant agreement 51NF40\_180545.}
\thanks{Manuscript received October XX, 2023.}
}

\begin{document}

\maketitle

\begin{abstract}
Autonomous Vehicles (AVs) promise a range of societal advantages, including broader access to mobility, reduced road accidents, and enhanced transportation efficiency. 
However, evaluating the risks linked to AVs is complex due to limited historical data and the swift progression of technology. 
This paper presents a data-driven framework for assessing the risk of different AVs' behaviors in various operational design domains (ODDs), based on counterfactual simulations of ``misbehaving" road users. 
We propose the notion of \emph{counterfactual safety margin}, which represents the minimum deviation from nominal behavior that could cause a collision. 
This methodology not only pinpoints the most critical scenarios but also quantifies the (relative) risk's frequency and severity concerning AVs.
Importantly, we show that our approach is applicable even when the AV's behavioral policy remains undisclosed, through worst- and best-case analyses, benefiting external entities like regulators and risk evaluators.
Our experimental outcomes demonstrate the correlation between the safety margin, the quality of the driving policy, and the ODD, shedding light on the relative risks of different AV providers. 
Overall, this work contributes to the safety assessment of AVs and addresses legislative and insurance concerns surrounding this burgeoning technology.
\end{abstract}

\section{Introduction}
\IEEEPARstart{A}{utonomous} Vehicles (\acrshortpl{av}) are poised to bring economic benefits, better accessibility to mobility, and an overall more efficient transportation system in the coming decades. More importantly, \acrshortpl{av} are expected to drastically reduce road accidents and thus actively save human lives.
Even today, every year, more than 1.35 million people die from road traffic accidents, with an additional 20-50 million injured, as reported by the World Health Organization~\footnote{\url{https://www.who.int/news-room/fact-sheets/detail/road-traffic-injuries}.}.

However, the question of whether \acrshortpl{av} are truly safer than traditional vehicles remains to be answered.
As shown in early studies by RAND Corporation~\cite{KalraDrivingReliability}, it is difficult to draw statistically sound conclusions about the real risk of \acrshortpl{av}. 
This is partly due to the limited amount of historical claim data currently available.
But also due to the ever-evolving software, hardware, and \acrshort{odd} of \glspl{av} that pose new challenges to traditional methods based heavily on historical data.
All of these factors present a significant challenge for tech developers, regulators, and insurance companies when it comes to assessing the risk of \acrshortpl{av}.

Currently, most of the safety assessments for \acrshortpl{av} are performed by the manufacturers themselves with little to no external evaluation from third parties--insurers and legal entities above others. 
Taking Waymo as an example, they have been transparent about their safety guidelines~\cite{Favaro2023BuildingRisk} and have conducted extensive tests on their vehicles that yielded encouraging results. 
By January 2023, they totaled 1 million rider-only miles with 2 major contact events and 18 minor ones~\cite{Victor2023SafetyMiles}.
They further showed that \acrshortpl{av} also have a bright future ahead in terms of handling emergency situations, as shown in~\cite{Scanlon2022CollisionCollisions}. 

However, in spite of these efforts, there is still one major open problem.
It is currently hard--if not impossible--for an external entity to quantify the actual risk associated to \acrshortpl{av}.
Moreover, regardless of how virtuously these vehicles can behave, there is no such a thing as ``zero risk'' on ``open'' roads with other humans.
A long tail of unfortunate events that are out of the \acrshort{av}'s control is statistically bound to happen.

In this work, we propose a safety evaluation framework based on counterfactual simulations.
We focus on the risk coming from other road users that ``misbehave''.
We show that the derived risk metrics could serve not only tech developers, but also external third parties that want to score the riskiness of different \acrshort{av}'s providers--even when the underlying driving policy is unknown.
The overwhelming importance of a risk factor based on others' misbehavior is also corroborated by the early results observed by Waymo, where the few contacts events observed in~\cite{Schwall2020WaymoData,Victor2023SafetyMiles} are ascribed to the misbehaviors of others. 

We address the problem in a data-driven way. Under the assumption of having access only to data collected by the \acrshort{av} during nominal operation and a set of counterfactual policies for the other agents. 
To this end, we introduce the concept of \emph{counterfactual safety margin}, defined as the minimal misbehavior by other road users that could have potentially led to a collision.
This metric is associated with an \gls{av}, but also with its \acrshortpl{odd}, naturally capturing all the external risk factors.
Moreover, we show that even without knowing the actual behavioral policy of the \acrshort{av}, one can perform a worst and best case analysis to provide an upper and lower bound on the safety margin of a certain pair of \acrshortpl{av} and \acrshortpl{odd}.

A persistent question that remains is identifying which counterfactuals to examine and determining their relative significance.
In our view, the creation of counterfactual policies should be informed by historical claim data from human-operated vehicles. 
Furthermore, \glspl{av} themselves offer readily available data that contain the common misbehavior of other road users in everyday scenarios, offering an additional and crucial source of information.
Together, these elements provide a broad and more nuanced perspective on the development of counterfactual policies, which is seen as a pivotal future avenue for research.

\subsection{Related Work}
\noindent Counterfactual analysis enables researchers to reason about ``what if" scenarios by comparing actual outcomes with hypothetical outcomes under alternative conditions. 
This approach is widely adopted in many scientific domains~\cite{Coston2020CounterfactualFairness}, most importantly, it is often crucial to understand the causal relationships between interventions (often referred to as ``treatments'') and outcomes~\cite{Pearl2000Causality:Inference}.
In medical research, for example, counterfactual analysis is widely used to study and evaluate the effectiveness of different treatments~\cite{Hofler2005CausalCounterfactuals}, in socioeconomics to judge the efficacy of a policy~\cite{Hicks1980CausalityEconomics}, or in engineering to compare different designs.

Recently, counterfactual reasoning has also gained popularity for the safety assessment of autonomous vehicles. 
Waymo, for example, showcased their vehicle behavior in specific test scenarios reconstructed from detailed police reports of fatal accidents~\cite{Scanlon2021WaymoDomain}. 
The \acrshort{av} was substituted in the simulation in different roles to demonstrate the possible effectiveness in mitigating, if not avoiding, damage.
The study was further developed comparing the avoidance capability of a human driver that is non-impaired, with eyes always on the conflict (NIEON model) to the Waymo vehicle~\cite{Scanlon2022CollisionCollisions}. 
Notably, the counterfactual paradigm is also used in the ``design phase'' of an \acrshort{av} stack to generate more heterogeneous training and testing scenarios~\cite{Igl2022Symphony:Simulation,ZhongGuidedSimulation,HartCounterfactualDriving,Nishiyama2020DiscoveringSimulation,VoloshinEventualReplay}.

Other attempts in assessing \acrshortpl{av}' riskiness often involve either first principle statistics~\cite{KalraDrivingReliability}, or reachability-based analysis~\cite{Lavaei2021FormalApproach,AlthoffReachabilityCars,Leung2021TowardsConcepts}. 
The latter, in particular, includes different approaches that engineer specific surrogate safety measures to gauge the criticality of a scenario. For instance, \cite{Schneider2021TowardsScenarios,Westhofen2023CriticalityArt} refine the standard notion of time-to-collision to account for the map topology, or the proposed \acrfull{rss} metrics could serve this purpose by measuring the violation rate of first principle physics-based safety distances\footnote{\url{https://www.mobileye.com/technology/responsibility-sensitive-safety/}.}.
In general, surrogate safety measures are computed by forward propagating the other agents according their motion model without any particular assumptions on their intentions and behavior. In contrast, it is at the core of our counterfactual approach the possibility to explicitly test against certain (mis)behaviors of others that one deems relevant. 

We envision that these counterfactual policies can indeed be learned and engineered based both on historical claim data but also on observed local cultural behaviors (e.g., Pittsburgh left). This naturally would also associate to each counterfactual a relevance weight.
In this regard, many developments have recently been proposed to create more \emph{controllable} simulations. An example is given by~\cite{ZhongGuidedSimulation}, which introduces a conditional diffusion model for traffic generation that allows users to control desired properties of the agents' trajectories.
Another example that aims to generate challenging scenarios is provided by~\cite{RempeGeneratingPrior} which introduces STRIVE. STRIVE is a method to automatically generate challenging scenarios that cause a given planner to produce undesirable behavior while maintaining the plausibility of the scenario.

\subsection{Statement of Contribution}
\noindent In this work, we introduce a framework for comparing the risk of different driving policies of \acrshortpl{av} operating in different \glspl{odd}.
The proposed risk assessment hinges on counterfactual simulations of what would have happened if others were to misbehave.
We introduce the concept of safety margin as the minimum counterfactual deviation that would cause a collision with non-negligible probability. Importantly, we consider counterfactuals to be parameterized by an intensity value which controls the degree of the counterfactual.

The counterfactual safety margin allows to automatically mine on large datasets without rare events the most critical scenarios. 
Furthermore, when a prior on the counterfactual likelihood is available, this risk measure can be related both to the frequency and the severity components of risk. In turn, this encourages an additional line of work from authorities and tech developers to provide statistical models of human misbehavior on the roads. 
Additionally, we demonstrate that even external entities without access to the \acrshort{av}'s behavioral policy can utilize the proposed methodology.
Indeed, one can provide a lower bound on the safety margin by assuming that the \acrshort{av} is non-reactive in the counterfactual simulation and an upper bound by computing what could have been the best possible reaction.  

Finally, we consider five possible counterfactual behaviors engineered from known common cases of human error.
We showcase experimentally that the safety margin definition is well-posed in the sense that higher counterfactual intensities lead to higher collision probability. 
Moreover, we show that the framework allows to naturally capture in a data-driven fashion also the degree of risk inherent in a specific \acrshort{odd}. To this end, we compare the results of the same type of vehicle driving on average in high-speed scenarios against slow-speed ones. 
Surprisingly, while the safety margin does not show a marked trend, its severity component clearly retrieves quantitatively the known correlation between speed and accidents' severity.
Finally, we show that the ``goodness'' of an \acrshort{av} policy is proportional to its safety margin. 
Namely, artificially synthesizing a policy that is strictly worse than another leads to a lower safety margin, hence higher risk.

To the best of our knowledge, this is one of the first works that provides an estimate of the \acrshort{av}'s risk from driving data that do not necessarily contain ``rare'' or ``surprising'' events.

\subsection{Organization of the manuscript}
\noindent In~\Cref{sec:sm} we introduce the main definitions and concepts related to the safety margin. 
In~\Cref{sec:sm_applied} we assume that the driving policy of the vehicle under scrutiny is not available and introduce upper and lower bounds of the safety margin metric.
Finally, in~\Cref{sec:exp} shows experimental results obtained on CommonRoad scenarios.
\begin{figure*}[ht!]
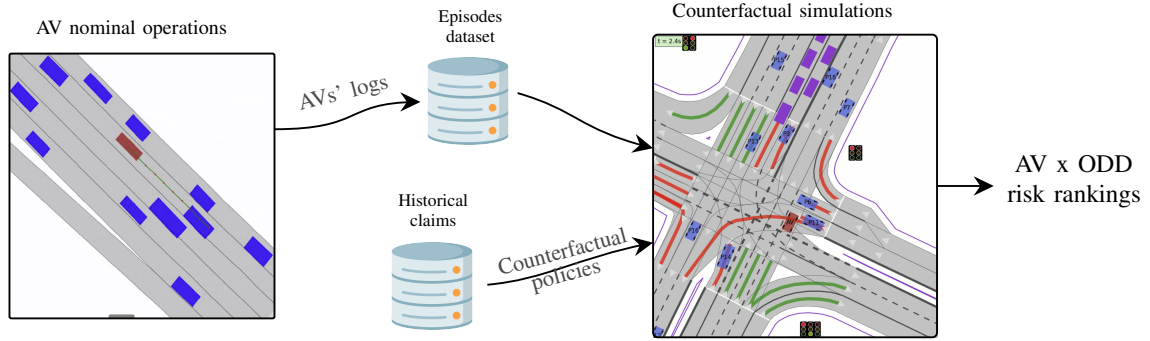

    \centering
	\includestandalone[width=.85\textwidth]{tikz/framework}
    \caption{The counterfactual safety margin provides a data-driven framework to score and compare the riskiness of \acrshort{av}'s providers operating in different \acrshort{odd}. 
    During nominal operations, an \acrshort{av} records \emph{episodes} (i.e., the state of the surrounding environment and of the other road users): these episodes represent the initial anchor in the counterfactual simulation. Given a set of counterfactual policies parametrized by a scalar intensity value, these are employed to re-simulate the episode with now the other agents behaving according to the counterfactual policy.
    The safety margin is then determined as the smallest counterfactual intensity for which a collision would have been ``likely to occur''. 
    The analysis can be carried out even without knowing the policy of the \acrshort{av} under scrutiny.
    }
    \label{fig:episode_example}
\end{figure*}
\section{Counterfactual Episodes and Safety Margin}\label{sec:sm}
\noindent The counterfactual analysis we propose has as starting point the data collected during nominal operations of an \gls{av}. These data must suffice to recreate a representation of what happened around the vehicle. To this end, we consider as required by the analysis a topological map of the road and the perceived state and occupancy of other relevant road agents (the ego vehicle, pedestrians, other vehicles, traffic lights, etc.). This collection of data over a certain time interval forms what we call an \emph{episode}. An episode is then utilized as the initial condition for the counterfactual analysis.

In fully generality, we consider a \emph{counterfactual} as a ``what-if'' scenario with respect to the original episode. But in this work we focus on a particular subclass of counterfactuals that can be parameterized by a scalar \emph{intensity value}.
The intensity value acts as a ``knob" that we can control to determine how much we are deviating from the original episode. 
At the high level, we want to evaluate what would have happened if another vehicle had had a low probability of not respecting the stop sign. What if the probability was higher? What would have happened if the driver behind us had been distracted looking at the phone when we braked. What if they were distracted a bit longer? What if the other vehicle had seen us only at a certain distance?

The main idea behind this work is to evaluate what is the maximum deviation--i.e., minimum counterfactual intensity--that the agent under scrutiny can tolerate without collision events. 
We name this particular quantity the \emph{counterfactual safety margin}. 
We observe that this quantity is determined by two main factors.
The first being the \gls{av}'s decision making itself, in particular via its resulting behavior on the road. The second is the \gls{odd} context. Certain environments result inherently to be more risky due to traffic conditions, local driving culture, and road infrastructure.

In the following we make these concepts more formal. 
We use game theoretic notation by denoting the quantities relative to the $i$-th agent with the $i$-th subscript, $-i$ for ``everyone but $i$'', and no subscript for the joint quantities relative to all agents. 
For instance, we denote with $\statespace_i$ the state space of the $i$-th agent. Its trajectory over a finite time horizon $T>0$ will then be $\state^{[0:\horizon]}_i$. 
Consequently, the trajectories of all agents will be $\state^{[0:\horizon]}$. These elements are already sufficient to define more formally an \emph{episode}.
\begin{definition}[Episode]~\label{def:episode}
An \emph{episode} $\episode$ consists of a finite set of agents $\agents$ and their relative trajectories over a finite interval of time $\horizon>0$. 
That is, $\episode \coloneqq \tup{\agents,\state^{[0:\horizon]}}$.
In probabilistic terms, we further consider an episode $\episode$ to be the realization of a random variable $\episoderv$.    
\end{definition}

An illustrative example is given by~\Cref{fig:episode_example}. Clearly there are infinitely many possible episodes, each representing a particular realization of the interaction among different agents on a specific road. Setting aside mathematical technicalities, we consider the space of episodes to be a probability space following the density function defined by the~\gls{odd}. This, in fact, defines the likelihood of experiencing certain episodes. This varies depending on many factors, most importantly the geographic region in which one drives determines how likely it is to drive a certain map topology and observe certain behaviors of the other traffic participants.
\begin{definition}[\acrfull{odd}]~\label{def:odd}
An \gls{odd} defines the probability density function of $\episoderv$, denoted as $\odd_{\episoderv}(\episode)$.
\end{definition}
Starting from an episode, we then consider a \emph{counterfactual} as a ``what-if'' scenario where the original episode is replayed--i.e., re-simulated--modifying the behavior of certain agents.
A \emph{counterfactual episode} is the realization of a simulation with the boundary initial conditions given by the original episode, hence an episode itself. 
In the counterfactual simulation each agent acts according to a certain policy that maps the state of the simulation to control commands of the agents. Allowing stochastic policies, we denote by $\policy_i: \statespace \to \simplex(\actionset_i)$ the policy for agent $i$. The simulator assumes a certain dynamic model for each agent such that the state evolution is determined by a discrete dynamic equation $\state(k+1) = \dynamics(\state(k),\ldots,\action_i(k),\ldots)$.
Allowing stochastic dynamics and policies the counterfactual itself becomes a distribution of episodes with respect to the original one, more formally: 
\begin{definition}[Counterfactual]~\label{def:counterfactual}
The \emph{counterfactual} of an episode $\episode$ is denoted as $\counterfactualrv(\episode)$ and it is fully determined by the tuple $\tup{\agents,(\policy^\counterfactualrv_i)_{i\in\agents},\dynamics,\state^0(\episode), \horizon}$. 
Again, $\counterfactualrv(\episode)$ shall be interpreted as a random variable whose realizations are (counterfactual) episodes~(\Cref{def:episode}). We denote by $\crealization (\episode)$ a specific realization of $\counterfactualrv(\episode)$.
\end{definition}
Akin to~\cite{Igl2022Symphony:Simulation}, a counterfactual episode can be seen as the realization of a Markov game, and $\counterfactualrv(\episode)$ is itself an episode.
We highlight that the policies can be constructed arbitrarily, for instance, they could come from learned behavioral models, fitted to the specific episode, or simply be constrained to reply the behavior observed in the original episode. 
From the original episode, for physics-based simulations, one may need to estimate the dynamics model of the different agents for realistic closed-loop simulation. However, we believe that standard car, truck, bicycle models shall suffice.  

While~\Cref{def:counterfactual} is quite general and allows to perform any type of ``what-if'' scenario, we focus on a particular subcategory for risk assessment.
In particular, we consider counterfactuals that can be parametrized by a scalar value $\cfintensity\in\reals_{\geq0}$ that we regard as the \emph{intensity} of the counterfactual.
For values close to zero we retrieve the original episode. Whereas by increasing the magnitude of the intensity, we simulate counterfactual that are ``further away'' from the original episode.   
Loosely speaking, one can interpret the counterfactual intensity as the magnitude of the introduced nuisance.

\begin{definition}[Scalar Counterfactual]~\label{def:scalar_count}
We call a \emph{scalar counterfactual} with intensity parameter $\cfintensity \in \reals_{\geq0}$ a counterfactual parametrized by $\cfintensity$ such that it is fully determined by $\tup{\agents,(\policy^\counterfactualrv_i(\cfintensity))_{i\in\agents}, \dynamics,\state^0,\horizon}$. We denote it as $\counterfactualrv(\episode,\cfintensity)$.
\end{definition}
Some of the scalar counterfactual examples with their respective intensity can include: 
\begin{itemize}
    \item What if they had slower reaction times? One can re-simulate an episode introducing latency in the observations received by the simulator, the magnitude of the latency represent the intensity of such a counterfactual. 
    \item What if they did not see me? The vehicle under scrutiny can be removed from the observations of the other agents unless closer than a certain threshold. The intensity is inversely proportional to this threshold. 
    \item What if they were distracted? For small periods of times other agents do not receive new observations. The intensity is given by the extension of the distraction period.
    \item What if they did not respect the stop sign? or the traffic light? We can introduce a probability associated to the binary decision of not respecting the traffic signs. The intensity is the probability itself in this case.
\end{itemize}
A more through description with the corresponding implementation will be made precise in \Cref{sec:exp}. 
\subsection{Counterfactual Safety Margin}
\noindent Given the scalar counterfactuals (\Cref{def:scalar_count}) we introduce the \emph{counterfactual safety margin} as the minimum intensity for which a contact event would occur. Where a contact event is a boolean function of a given episode realization. In practice, this simply amounts to performing collision checking on the agents' trajectories.  
\begin{definition}[Contact Event]~\label{def:collision}
A contact event for the $i$-th agent in the $\episode$ episode, is the realization of a function $\coll_i: \episode \mapsto \bool$. Returns \textsc{True} if the $i$-th agent collided in the episode, \textsc{False} otherwise. 
\end{definition}
Note that the notation introduced up to this point has the following implications:
\begin{itemize}
    \item $\coll_i(\episode) \in \bool$: given an episode, it tells us if a collision happened.
    \item $\coll_i(\counterfactualrv(\episode, \cfintensity)) \sim \bernoulli(\param)$: given a counterfactual, its realization can be stochastic, $\param$ is in general unknown and can only be estimated be (re-)simulating the counterfactual episode many times. If the counterfactual simulation is fully deterministic (policies, parameters, and simulator), then $\param \in \{0,1\}$.
\end{itemize}
Hence, we express the minimum intensity that would cause a collision as follows:
\begin{definition}[Counterfactual Safety Margin]~\label{def:cf_safetymargin}
Let $\counterfactualrv$ to be a scalar counterfactual parametrized by $\cfintensity \in [0,\cfintensity_{\max}]$ (\Cref{def:scalar_count}) and $\episode$ be an episode. 
We define the \emph{counterfactual safety margin} for the $i$-th agent as the smallest intensity $\safetym$  which causes Agent $i$ to collide with a non-negligible probability $\epsilon$:  
\begin{equation}\label{eq:safety_margin}
    \safetym_i (\counterfactualrv(\episode)) 
  \begin{cases}
  \in \argmin_{\cfintensity \in [0,\cfintensity_{\max}]} &\textit{s.t. } \prob \left[ \coll_i(\counterfactualrv(\episode, \cfintensity))\right] >\epsilon ;\\
  >\cfintensity_{\max} &\text{otherwise}.
  \end{cases}
\end{equation}
\end{definition}
\noindent Some observations:
\begin{itemize}
    \item A small safety margin implies a higher risk for the agent since a smaller counterfactual deviation would result in a collision.
    \item An episode is analysed in the proposed counterfactual framework by simulating different counterfactual intensities. 
    The main insights are then obtained by plotting the safety margin curves shown in~\Cref{fig:counterfactual_sm}, where the collision probability is plotted as a function of the counterfactual intensity.
    \item $\epsilon$ is arbitrary. Its role is to threshold a certain significance level for the collision probability removing the sensitivity to rare stochastic realizations in the counterfactual simulations. Notice in~\Cref{fig:counterfactual_sm} that for deterministic frameworks its value is irrelevant. Moreover, also in the stochastic framework of our experiments, we usually observed sharp increases in the collision probability around a certain intensity, making the specific choice of $\epsilon$ irrelevant.
    \item In many cases the vehicle might result to be insensitive to the specific counterfactual. Either because of the specific episode setup or because the real value falls beyond the range that has been tested. 
    We assign to these cases a special value ($>\cfintensity_{\max}$). 
\end{itemize}
\begin{figure}[htb]
\centering
	\begin{subfigure}[b]{.4\columnwidth}
		\centering
			\tikzset{every picture/.style={line width=0.75pt}} %

\begin{tikzpicture}[x=0.75pt,y=0.75pt,yscale=-1,xscale=1]
\draw    (40.1,94.73) -- (40,18) ;
\draw [shift={(40,15)}, rotate = 89.93] [fill={rgb, 255:red, 0; green, 0; blue, 0 }  ][line width=0.08]  [draw opacity=0] (10.72,-5.15) -- (0,0) -- (10.72,5.15) -- (7.12,0) -- cycle    ;
\draw    (35.6,89.98) -- (167,90) ;
\draw [shift={(170,90)}, rotate = 180.01] [fill={rgb, 255:red, 0; green, 0; blue, 0 }  ][line width=0.08]  [draw opacity=0] (10.72,-5.15) -- (0,0) -- (10.72,5.15) -- (7.12,0) -- cycle    ;
\draw  [dash pattern={on 0.84pt off 2.51pt}]  (40,40) -- (150,40) ;
\draw [color={rgb, 255:red, 208; green, 2; blue, 27 }  ,draw opacity=1 ] [dash pattern={on 4.5pt off 4.5pt}]  (40,80) -- (150.07,80.2) ;
\draw    (150.07,86.87) -- (150,92.53) ;
\draw    (109.92,87.12) -- (109.92,92.78) ;
\draw    (110,90) -- (110,40) ;
\draw    (150,40) -- (110,40) ;

\draw (27,30.5) node [anchor=north west][inner sep=0.75pt]   [align=left] {1};
\draw (142.33,93.9) node [anchor=north west][inner sep=0.75pt]    {$\cfintensity _{\max}$};
\draw (45.5,6) node [anchor=north west][inner sep=0.75pt]    {$\mathbb{P}[\mathrm{coll}_{i}(\mathsf{C}( e,\cfintensity ))]$};
\draw (28.33,92) node [anchor=north west][inner sep=0.75pt]   [align=left] {0};
\draw (28,69.33) node [anchor=north west][inner sep=0.75pt]   [align=left] {$\displaystyle \textcolor[rgb]{0.82,0.01,0.11}{\epsilon }$};
\draw (102,93.4) node [anchor=north west][inner sep=0.75pt]    {$\cfintensity ^{*}$};
\draw (170.33,73.23) node [anchor=north west][inner sep=0.75pt]    {$\cfintensity $};

\end{tikzpicture}
		\caption{Deterministic case.}
		\label{subfig:counter_stoc}
	\end{subfigure}
	~
	\begin{subfigure}[b]{.4\columnwidth}
		\centering
			\tikzset{every picture/.style={line width=0.75pt}} %

\begin{tikzpicture}[x=0.75pt,y=0.75pt,yscale=-1,xscale=1]
\draw    (40.1,94.73) -- (40,18) ;
\draw [shift={(40,15)}, rotate = 89.93] [fill={rgb, 255:red, 0; green, 0; blue, 0 }  ][line width=0.08]  [draw opacity=0] (10.72,-5.15) -- (0,0) -- (10.72,5.15) -- (7.12,0) -- cycle    ;
\draw    (35.6,89.98) -- (167,90) ;
\draw [shift={(170,90)}, rotate = 180.01] [fill={rgb, 255:red, 0; green, 0; blue, 0 }  ][line width=0.08]  [draw opacity=0] (10.72,-5.15) -- (0,0) -- (10.72,5.15) -- (7.12,0) -- cycle    ;
\draw    (40,90) .. controls (149,91) and (103.4,58.2) .. (150.07,46.87) ;
\draw  [dash pattern={on 0.84pt off 2.51pt}]  (40,40) -- (150,40) ;
\draw [color={rgb, 255:red, 208; green, 2; blue, 27 }  ,draw opacity=1 ] [dash pattern={on 4.5pt off 4.5pt}]  (40,80) -- (150.07,80.2) ;
\draw    (150.07,86.87) -- (150,92.53) ;
\draw    (109.92,87.12) -- (109.92,92.78) ;

\draw (27,30.5) node [anchor=north west][inner sep=0.75pt]   [align=left] {1};
\draw (142.33,93.9) node [anchor=north west][inner sep=0.75pt]    {$\cfintensity _{\max}$};
\draw (45.5,6) node [anchor=north west][inner sep=0.75pt]    {$\mathbb{P}[\mathrm{coll}_{i}(\mathsf{C}( e,\cfintensity ))]$};
\draw (28.33,92) node [anchor=north west][inner sep=0.75pt]   [align=left] {0};
\draw (28,69.33) node [anchor=north west][inner sep=0.75pt]   [align=left] {$\displaystyle \textcolor[rgb]{0.82,0.01,0.11}{\epsilon }$};
\draw (102,93.4) node [anchor=north west][inner sep=0.75pt]    {$\cfintensity ^{*}$};
\draw (170.33,73.23) node [anchor=north west][inner sep=0.75pt]    {$\cfintensity $};

\end{tikzpicture}
		\caption{Stochastic case.}
		\label{subfig:counter_det}
	\end{subfigure}
    \caption{The counterfactual analysis aims to find to smallest intensity for which the probability of collision surpasses a certain threshold $\epsilon$.}
    \label{fig:counterfactual_sm}
\end{figure}
\subsection{Averaging over an \acrshort{odd}}
\noindent When evaluating an agent, the analysis will be carried out on a dataset of episodes.
The first straightforward result will be a list of episodes that are more risky--the ones with the lowest safety margin.
This information is relevant both for tech developers as well as for external regulatory entities.
Importantly, we distinguish two cases depending on whether any prior on the counterfactual likelihood is available.

If no counterfactual likelihood is available, the framework allows one to naturally recognize the scenarios that are potentially more critical. Furthermore, it allows to compare (i.e., provide an order) to different pairs of agent and \acrshort{odd}.
Indeed analogous curves to the stochastic case of~\Cref{fig:counterfactual_sm} are obtained by averaging over the \acrshort{odd}. Clearly the analysis can go more in depth taking into account other statistics, for instance, the frequency and the severity of the counterfactuals that have a low safety margin.

When the likelihood of a counterfactual scenario is available it is possible to weight the importance of each counterfactual simulation.
For instance, we can expect to have statistics that serve as proxies for each a counterfactual likelihood.
An example could be given by knowing how often people run over a red light or other infraction occur~\cite{Stewart2022Overview2020}.
Most importantly, this information would allow to better relate the safety margin curve of an agent to the frequency component of risk.

\section{Upper and Lower Bounds for Counterfactual Safety Margin}\label{sec:sm_applied}
\begin{figure}[htb!]
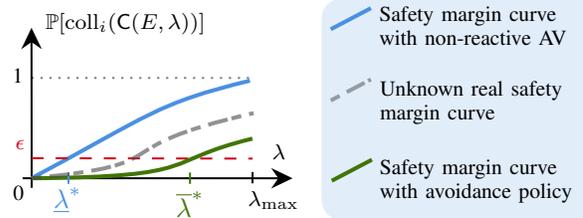

    \centering
        \includestandalone[width=.45\textwidth]{tikz/sm_margins}
    \caption{Computable safety margin's bounds under the assumption of not knowing the $\av$'s behavioral policy. 
    }
    \label{fig:sm_bounds}
\end{figure}

\noindent
So far, to perform a counterfactual simulation one would also need the behavioral model of the \acrshort{av}.
While this is the case for \acrshort{av} companies, it might not be the case for external entities.  
Therefore, in the following we assume that we do not know the policy of the \acrshort{av} under scrutiny.
In this setting, the external evaluator still has access to a significant collection of episodes recorded during nominal operation of the~\gls{av}, but the reactive policy (i.e., the decision making model) of the~\gls{av} is not available.
This assumption finds ground in the real world where, very likely, an \gls{av} provider is not willing to share externally its vehicles' behavioral model. 
\begin{assumption}\label{ass:avunknown}
    Let the vehicle of interest be $\av\in\agents$. The policy $\policy_{\av}(\cdot)$ is unknown.
\end{assumption} 
In the following, we show that despite~\Cref{ass:avunknown}, all the methodology introduced so far can still be applied.
In particular, one can consider two cases in the counterfactual analysis:
\begin{enumerate}[label=(\roman*)]
    \item The \gls{av} is \textbf{non-reactive} and replays the trajectory of the original episode independently of what the others do;
    \item The \gls{av} is \textbf{omniscient} and behaves in order to obtain the best possible outcome in the counterfactual simulation. 
\end{enumerate}
Note that (i) amounts to a simulation with the vehicle of interest non-reactive, meaning that it will replay the initial episode trajectory independently of the others' counterfactual behavior.
(ii) boils down to a single-agent optimal control problem over a finite horizon. 
The result of such analysis will provide two safety margin lines plot as shown in~\Cref{fig:sm_bounds}.  

\subsubsection{Lower Bound - Non-reactive AV}
This is the simpler to compute. 
In the counterfactual simulation, the $\av$ replays its original trajectory irrespective of what the others' do.
This outcome represents a lower bound for the safety margin, assuming that the \acrshort{av} would have taken action to mitigate the situation, surpassing a `do-nothing' response.

\subsubsection{Upper Bound - Best-case Outcome}
(ii) requires a more nuanced explanation.
First, we anchor the notion of \emph{best outcome} to the collision happening in the non-reactive case (i) and its relative emergency maneuver.
This is done to avoid a pointless discussions of the type: <<The best outcome could have been achieved by not having taken that road in the first place>>. 

Starting from the initial condition of the episode, we would like to find a trajectory for the $\av$ which solves the following optimal control problem:
\begin{equation}\label{eq:ocp}
\begin{aligned}
        \min_{\action_\av^{[0:\horizon]}}\quad & \sum_{k=0}^\horizon\collseverity_\av(\state(k)) \\
        \text{subject to}\quad & \state(k+1) = \dynamics(\state(k),\action_\av(k),\policy_{-\av}^{\counterfactualrv}(\state(k)) ) \\
		& \state(0)=\state^0.
\end{aligned}
\end{equation}
Note that \eqref{eq:ocp} can be seen as a single-agent optimal control problem.
Nevertheless, it is computationally hard to solve. 
First, due to the presence of other agents, the state space is highly dimensional, and the dynamics of the system is governed also by the others' input. Second, the counterfactual policies of others might be known only in terms of input-output relation (i.e., black-box models).
Third, the real cost function that we want to minimize is discontinuous with respect to contact events. Moreover, it includes damage models that are highly non-linear and non-convex.

In this work we consider a cost function $\collseverity$ modeled after the standard \acrlong{mais}\cite{States1969TheScales}.
More specifically we compute the damages deriving from a collision according to the model proposed in~\cite{Malliaris1997} which computes the probability of experiencing a certain degree of injury severity given the velocities, and the angle and point of impact.
Here we recognize another useful direction of research to develop more accurate injury models taking into account the occupancy of the vehicle and where the passengers are seating.    
In our case, we lexicographically minimize respectively the probability of a \textsc{Fatality}, a \textsc{MAIS3+} injury, and of a \textsc{MAIS2+} injury. 

\begin{table*}[htb!]
    \footnotesize
	\centering
	\begin{threeparttable}
		\setlength\tabcolsep{0pt} %
		\captionsetup{labelsep=period, skip=5pt}
		\caption{}
		\label{tab:counterfactuals}
		\begin{tabular*}{\textwidth}{l@{\extracolsep{\fill}}*{6}{c}}
			\toprule
			     & \multicolumn{5}{c}{\textbf{Counterfactual}}  \\
			\cmidrule{2-6} 
			&  Aggressiveness & Distraction & Illegal precedence & Impaired reflexes & Unseen  \\ \midrule
			\textbf{\makecell[l]{Counterfactual\\intensity}}  & \makecell[l]{Aggressiveness\\parameter of\\IDM policy} & \makecell[l]{Drop period\\of agents'\\observations} & \makecell[l]{Prob. of an agent\\not respecting\\a precedence signal\\(traffic sign/light)} & \makecell[l]{Latency added to\\the agents'\\observations}  & \makecell[l]{Inverse of the\\distance at which\\other agents' see\\the $\av$}         \\
                \textbf{Intensity range}  & [0,1] & [0,5]$(s)$& [0,1]& [0,1]& [0,20]$(1/m)$\\
			\textbf{Policy agnostic}  & \text{No} & \text{Yes} & \text{Yes} & \text{No}& \text{Yes}        \\
			\bottomrule
		\end{tabular*}
	\end{threeparttable}
\end{table*}
\section{Experiments}\label{sec:exp}
\noindent
We showcase the presented methodology on reproducible, publicly available CommonRoad scenarios~\cite{Althoff2017}.
In these experiments, we begin with a common initial step: Given a CommonRoad scenario, we first run a simulation to generate a synthetic episode.
This mimics the data that an \acrshort{av} would record during nominal operations. 
One vehicle gets to play the role of the $\av$ and logs all the other visible agents by simulating a planar 2D laser. Subsequently, the recorded log (i.e., the episode) is used to build the counterfactuals.

\begin{figure}[htb!]
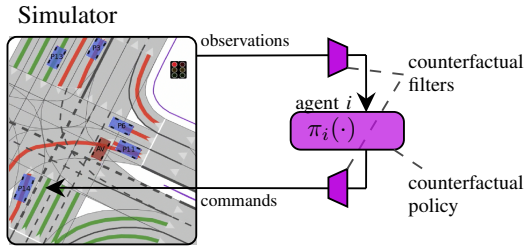

	\centering
	\includestandalone[width=.8\columnwidth]{tikz/sim}
	\caption{Each agent receives its own observations from the simulator, comprising of the state and the occupancy of the nearby agents.
	These are fed to the policy which is expected to return commands (acceleration and steering derivative) that are used to update the corresponding physical model in the simulation.
	A counterfactual policy can be obtained acting only on the observations and commands (purple filters), or by modifying the policy itself.
    The former method allows to be \emph{agnostic} of the original policy, thus it can be easily integrated with black-box models of learned policies. This allows the original policy to be of any type (model-based, learned,\dots).
    The latter relies on a particular parameterization of the original policy to modify its behavior.
    For example, increasing the aggressiveness parameter of an \gls{idm}.
	}
	\label{fig:sim}
\end{figure}

Given an episode, the counterfactual episode is then simulated by taking the initial condition of the log and spawning another simulation where each agent behaves according to the designed counterfactual policy.
Each simulation is carried out with the agent-based simulator provided by~\cite{dgcommons}.
As shown in~\Cref{fig:sim}, the policy of each agent is clearly separated from the rest.
The simulator generates observations for each agent in the form of perceived occupancy and state of the other surrounding agents, these are fed to the agent's policy that returns control commands (acceleration and steering) that are fed back to the simulator.
\subsection{Counterfactual details}
\noindent We implemented and considered the counterfactuals of~\Cref{tab:counterfactuals}.
\paragraph*{Aggressiveness} This counterfactual corresponds to the other drivers being more ``aggressive''.
Among the numerous way in which one could parameterize this driving style, we use a generalized \gls{idm} parameterization of~\cite{Kreutz2021AnalysisIntersections} and increase the corresponding ``aggressiveness'' parameter.
\paragraph*{Distraction} What if the others were distracted behind the wheel?
We mimic a driver getting distracted by not updating its observations for relatively short period of time--as if they were to look away from the road.
This counterfactual intensity corresponds to the distraction period. After each ``distracted'' period we have an ``attentive'' period of a fixed duration ($0.5\,s$).
\paragraph*{Illegal precedence} When an \gls{idm} agent encounters a stop sign or a red light this acts as a planning constraint bringing them to an halt.
We randomly draw whether the agent will ignore this constraint. 
The counterfactual intensity is the probability of the agent ignoring this precedence rule.
As for the aggressiveness counterfactual, this is currently implemented relying on the model-based \gls{idm} policy. 
Nevertheless, we imagine that also policy-agnostic implementations are possible if the traffic sign observation are passed through the ``hallucinated observations''.
\paragraph*{Impaired reflexes} In this policy-agnostic counterfactual the observations of the simulator are delayed to the agent. 
The introduced latency aims to mimic the driving behavior of a person with impaired reflexes or under the influence of substances. On a behavioral level it translates to slower reaction times an more ``wiggly'' behaviors such as poor lane keeping.  
\paragraph*{Unseen} What if the others had not seen us? In this counterfactual we remove from the others' observations our presence until a certain distance. At smaller counterfactual intensities others will see us from distance, whereas, as we increase the intensity, they will see us only in close proximity.

\subsection{Validation Method}
\noindent %
We validate the proposed counterfactual framework with the following experiments:
\begin{enumerate}[label=E.\arabic*]
    \item \label{exp:monotone}First, we verify that the safety margin curves are monotonically increasing with respect to the counterfactual intensity. 
    That is, as we increase the counterfactual intensity, \emph{on average}, also the probability of collision increases. 
    This monotonicity verifies that the minimization problem~\eqref{eq:safety_margin} is meaningful.
    \item \label{exp:odds}We test whether the method is well suited for ranking $\av\times$\gls{odd} pairs.
        We fix the agent and compare different \glspl{odd}. Namely, we artificially separate the scenarios in two \gls{odd} categories: \highspeed and \lowspeed. We show that since the proposed method is fully data-driven, it does not require any type of \gls{odd} labeling and classification, the resulting differences in risk of driving in different \glspl{odd} are directly reflected in the safety margin analysis.
    \item \label{exp:agents}This time we fix the \gls{odd} and compare different \acrshortpl{av}.
    In order to validate our method, we compare agents that are \emph{by construction} one better than the other.
        We achieve this by degrading the policy of the other agents under scrutiny by applying the counterfactuals to their nominal policy. For example, we consider a nominal \gls{idm} agent, the same agent that introduces a small delay, and the same agent with a limited field of view.    
        We then compare the resulting behaviors in the same counterfactual way as for the others. 
        As shown in~\cref{fig:agents_comp}, we find that the degraded policies result in behaviors that have a lower safety margin.
\end{enumerate}

\begin{figure*}[ht]
    \centering
    \begin{subfigure}{0.33\linewidth}
        \centering
        \includegraphics[width=\linewidth]{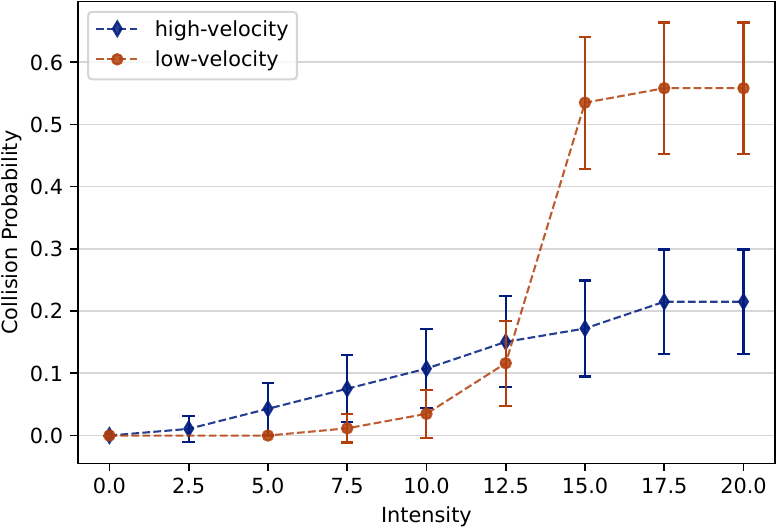}
        \caption{Unseen.}
    \end{subfigure}
    \begin{subfigure}{0.33\linewidth}
        \centering
        \includegraphics[width=\linewidth]{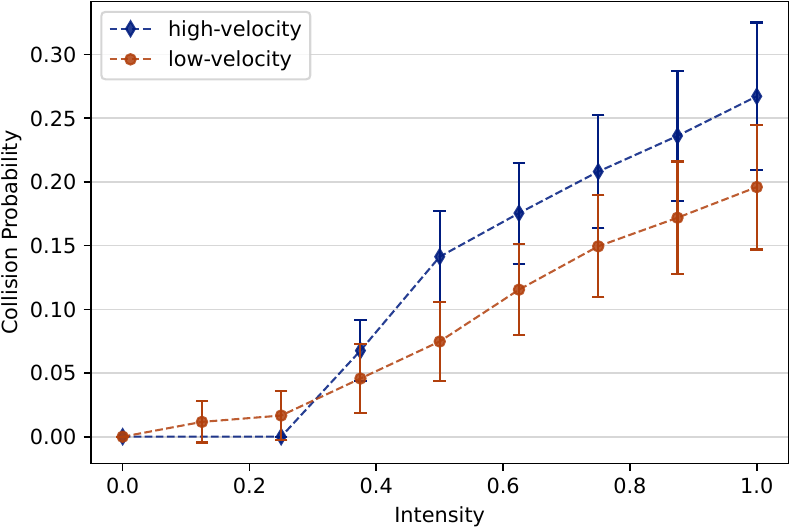}
        \caption{Distraction.}
    \end{subfigure}
    \begin{subfigure}{0.33\linewidth}
        \centering
        \includegraphics[width=\linewidth]{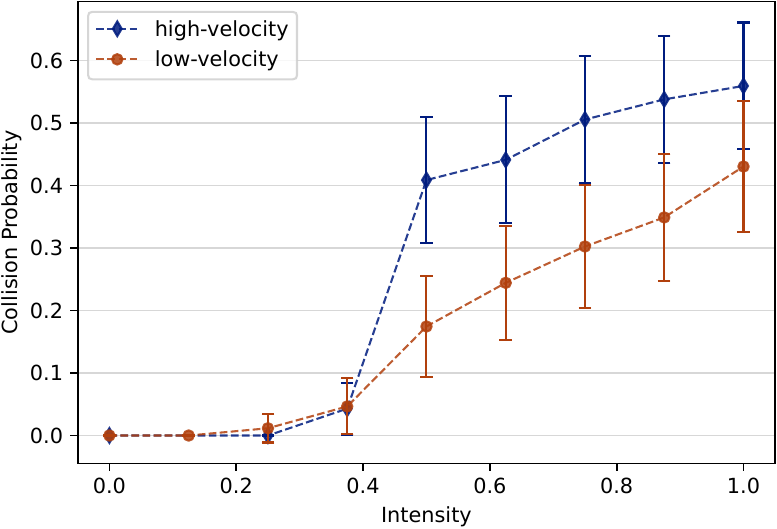}
        \caption{Impared reflexes.}
    \end{subfigure}
    \\
    \begin{subfigure}{0.33\linewidth}
        \centering
        \includegraphics[width=\linewidth]{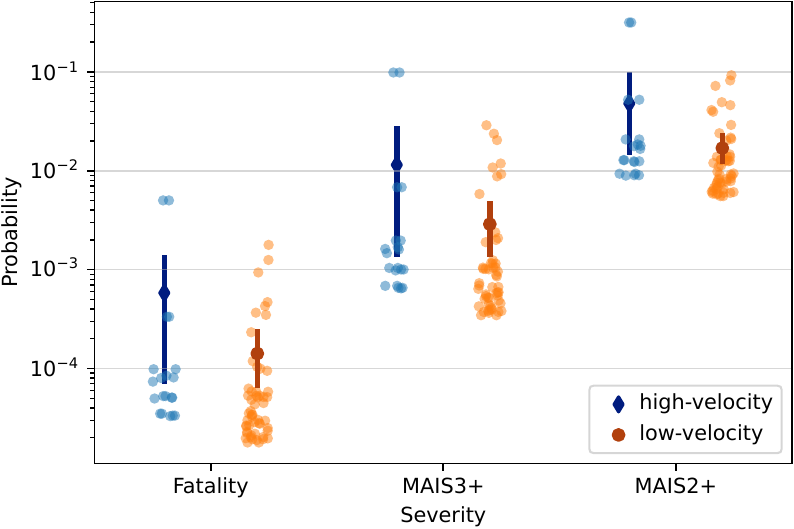}
        \caption{Unseen.}
    \end{subfigure}
    \begin{subfigure}{0.33\linewidth}
        \centering
        \includegraphics[width=\linewidth]{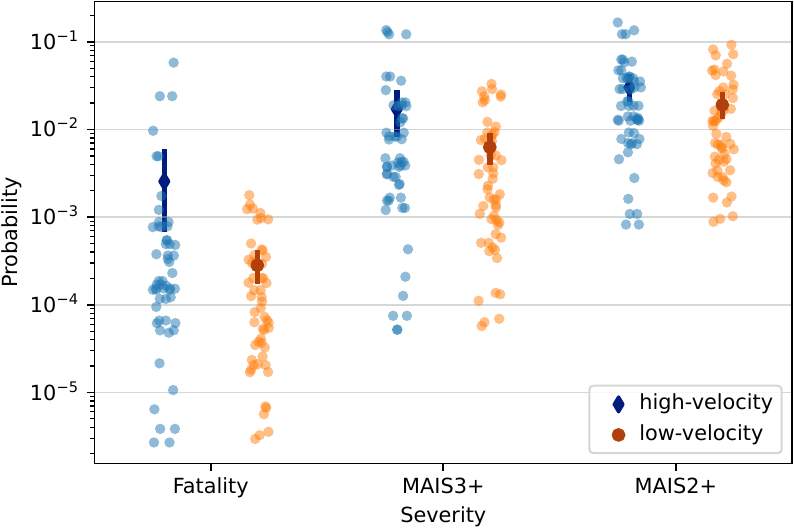}
        \caption{Distraction.}
    \end{subfigure}
    \begin{subfigure}{0.33\linewidth}
        \centering
        \includegraphics[width=\linewidth]{figures/severity_odd/SafetyMarginSeverity-Comparison-DistractedCounter-SafetyMarginSeverity.pdf}
        \caption{Impared reflexes.}
    \end{subfigure}
    \caption{The framework we have proposed allows us to quantitatively reaffirm an already established notion: speed is directly related to the severity of an event. This serves not just as a validation of our methodology, but crucially, demonstrates that our approach can measure the intrinsic risk associated with a particular \acrshort{odd}. The plots show the \emph{severity} information computed at the safety margin intensity. These, according to the damage model provided in~\cite{Malliaris1997}, map a collision dynamics to the likelihood of experiencing a particular degree of damage severity. Depicted with a bold stroke are the average and the 95\% CI. }
    \label{fig:severity_odd}
\end{figure*}

\subsubsection*{\labelcref{exp:monotone} - Safety margin curve monotonicity}
We experienced that the safety margin curves for the proposed counterfactuals satisfied \emph{on average} monotonicity.
Clearly, given the high non-linearity of the collision dynamics and the policy perturbations, this cannot be guaranteed for every single instance of the counterfactual episodes. While we observed some cases in which a very high counterfactual disturbance would actually avoid a collision, this is not true on average. 
This sanity check reassures us that the the safety margin minimization problem of \Cref{def:cf_safetymargin} is well-posed and meaningful. 
In the first row of~\Cref{fig:severity_odd} we report the results from simulating and evaluating an agent on 100 episodes against three different counterfactuals.
\subsubsection*{\labelcref{exp:odds} - Same agent, different \glspl{odd}}
We show that the proposed methodology naturally accounts for the different external risk factors coming from the \acrshort{odd}.
To this end, we evaluate the same agent type on two different sets of scenarios (i.e., \acrshort{odd}): \highspeed and \lowspeed.
For each set, we evaluate 100 episodes. 
The episodes are differentiated based on whether the average initial velocity of the agents exceeds or is less than $12$ \textit{m/s} ($\sim40$ \textit{km/h}).

Interestingly, we show that the proposed framework naturally provides the possibility to compute not only the safety margin but also its corresponding severity. 
This allows performing a more nuanced analysis that better captures the two main components of risk: \emph{frequency} and \emph{severity}. 
Specific results are shown in~\Cref{fig:severity_odd}.
\subsubsection*{\labelcref{exp:agents} - Same \gls{odd}, different agents}
\begin{figure*}[ht]
    \centering
    \begin{subfigure}{0.33\linewidth}
        \centering
        \includegraphics[width=\linewidth]{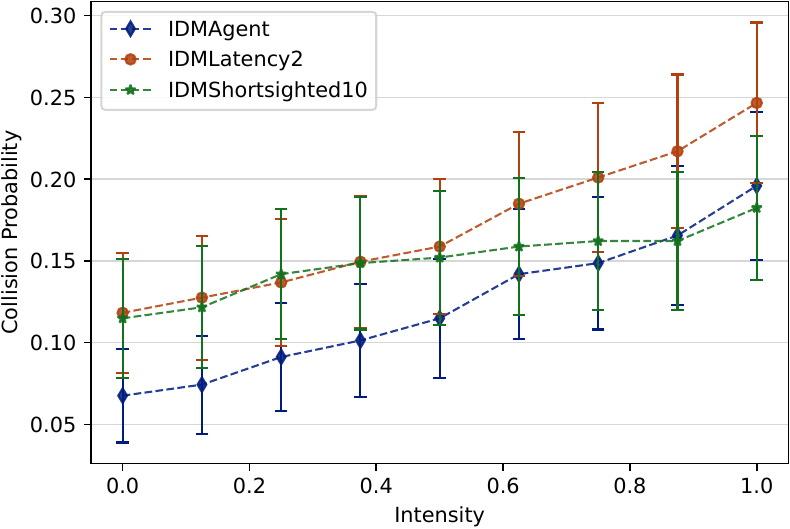}
        \caption{Aggressiveness.}
    \end{subfigure}
    \begin{subfigure}{0.33\linewidth}
        \centering
        \includegraphics[width=\linewidth]{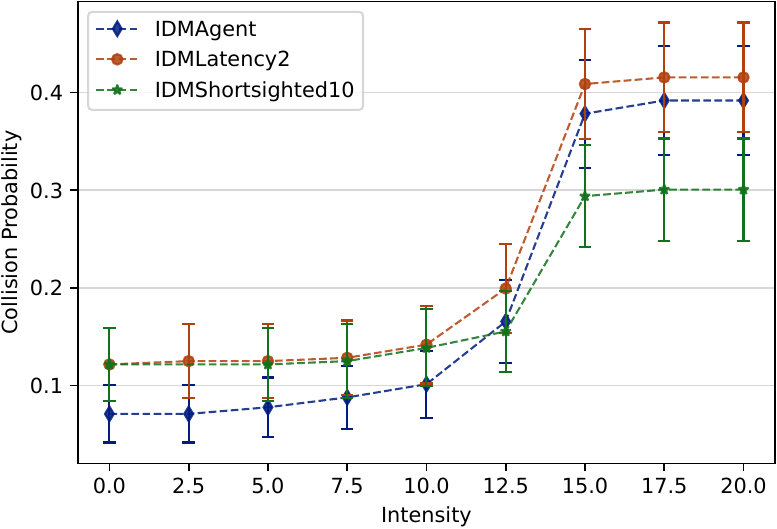}
        \caption{Unseen.}
    \end{subfigure}
    \begin{subfigure}{0.33\linewidth}
        \centering
        \includegraphics[width=\linewidth]{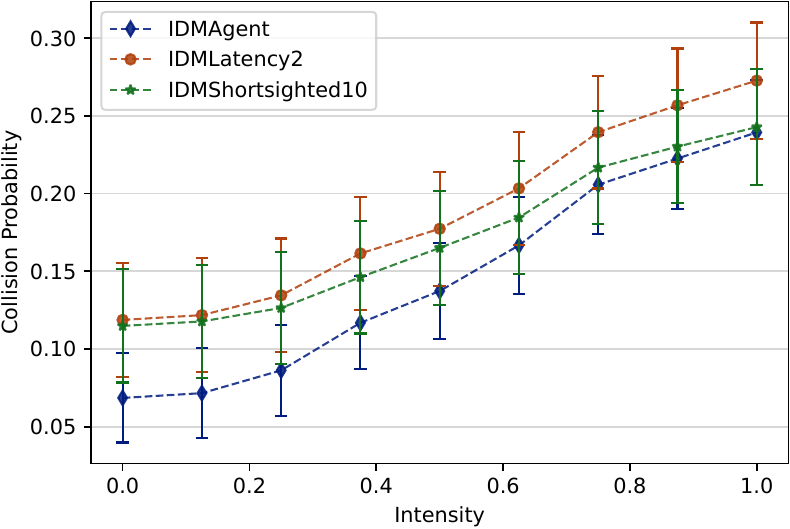}
        \caption{Distraction.}
    \end{subfigure}
    \\
    \begin{subfigure}{0.33\linewidth}
        \centering
        \includegraphics[width=\linewidth]{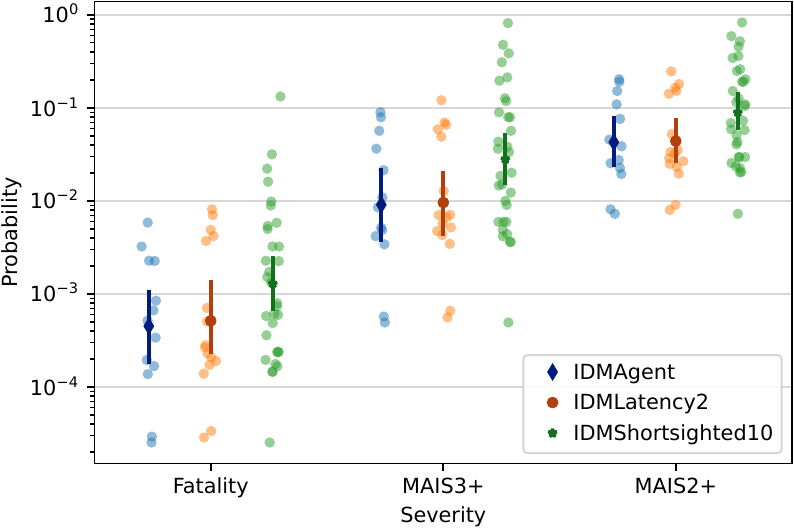}
        \caption{Aggressiveness.}
    \end{subfigure}
    \begin{subfigure}{0.33\linewidth}
        \centering
        \includegraphics[width=\linewidth]{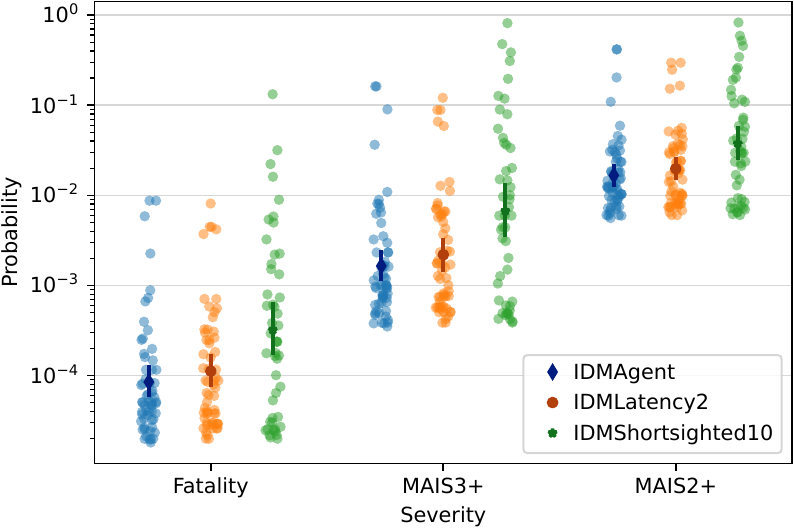}
        \caption{Unseen.}
    \end{subfigure}
    \begin{subfigure}{0.33\linewidth}
        \centering
        \includegraphics[width=\linewidth]{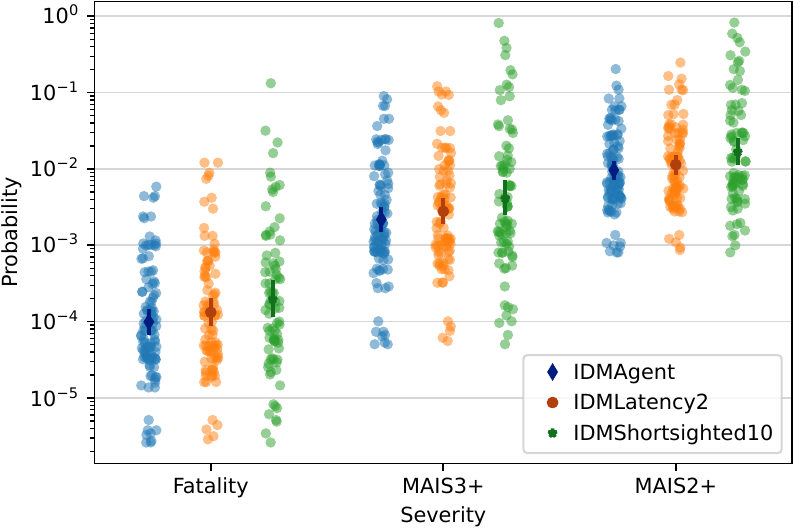}
        \caption{Distraction.}
    \end{subfigure}
    \caption{We compare three different agents operating in the same \gls{odd}. We artificially degrade a nominal \textsc{IDMAgent} policy by introducing a small latency (\textsc{IDMLatency2}) and limited field of view (\textsc{IDMShortsighted}). In the first row of the plots we can appreciate that the blue line appears quite consistently below the others, signifying a higher safety margin.
    The severity graphs instead do not show a significant difference across the agents. In conjunction with~\cref{fig:severity_odd} these seem to suggest that the severity is mainly determined by contextual factors (i.e., the \gls{odd}).}
    \label{fig:agents_comp}
\end{figure*}

Finally, we compare different types of agents operating within the same \gls{odd}.
By deliberately impairing the performance of a standard \gls{idm} driver using the counterfactuals directly, we derive \textsc{IDMlatency2}, introducing a $.2$ \textit{s} observation delay, and \textsc{IDMShortsighted10}, which discards observations beyond a distance of $10$ meters.
While the nominal base policy yields a broadly analogous behavior, the introduced nuisances cause the compromised agents to exhibit more unsafe practices, such as late braking and poorer lane keeping.
We then examine if this ranking is consistent with safety margins, with comprehensive results displayed in \Cref{fig:agents_comp}. As illustrated in the top row of~\Cref{fig:agents_comp}, a single-sided decline in an agent's performance inversely correlates with the safety margin value.

\subsection{Discussion}
\noindent The presented framework facilitates a comparative analysis of the behaviors exhibited by different autonomous vehicles while operating in certain \acrshortpl{odd}.
In particular, we initiated our study using data that may not necessarily include unfortunate rare events, such as collisions. Nevertheless, we adopt a data-driven approach to quantify the safety margin of each vehicle concerning potential counterfactual misbehavior of the other agents.\looseness-1

It is important to note that our evaluation is primarily focused on the resulting behavior from a phenomenological perspective. Consequently, behaviors leading to close calls and reduced safety margins could arise due to shortcomings in the vehicle's perception, planning, or control systems, as well as from the attributes of the surrounding environment—the \acrshort{odd} within which it operates. An example is provided in~\Cref{fig:severity_odd}. 

While our framework provides a mean to score behaviors based on certain counterfactual policies, an essential avenue for future research lies in developing such counterfactual policies. 
Specifically, we recognize the significance of leveraging a combination of historical data claims with the observed misbehavior of other agents on public roads to synthesize relevant counterfactual policies. These policies should encompass the most common human errors, enabling us to derive safety margin scores that strongly correlate with real-world risk.
Moreover, we emphasize the necessity of subjecting the presented framework to a rigorous statistical treatment to establish the confidence associated with the results that one may derive from a dataset.

\section{Conclusions}
In conclusion, our proposed framework offers a comprehensive approach to comparing and evaluating the behaviors of autonomous vehicles in different \acrshortpl{odd}. 
The integration of counterfactual analysis and statistical treatment will play a crucial role in ensuring the accuracy and practical applicability of our safety margin scores. This research contributes to the advancement of autonomous vehicle technology and its safe deployment in real-world scenarios.
Importantly, this methodology is suited for adoption by various stakeholders, including \gls{av} suppliers, as well as third-party entities such as insurance companies and regulators.

On a more technical side, this method opens up also the inverse question for tech developers. What are driving behaviors and policies that maximize the counterfactual safety margin?
\section{Acknowledgement}
\noindent
The authors thank Shuhan He for the fruitful discussions and the help with the behavioral models implemented in the simulator.

\bibliographystyle{./IEEEtrannourl}
\bibliography{IEEEabrv, references}

\end{document}

\tikzset{every picture/.style={line width=0.75pt}} %

\begin{tikzpicture}[x=0.75pt,y=0.75pt,yscale=-1,xscale=1]
\draw  [draw opacity=0][fill={rgb, 255:red, 74; green, 144; blue, 226 }  ,fill opacity=0.16 ] (184.5,10.25) .. controls (184.5,4.86) and (188.86,0.5) .. (194.25,0.5) -- (319.25,0.5) .. controls (324.64,0.5) and (329,4.86) .. (329,10.25) -- (329,101) .. controls (329,106.39) and (324.64,110.75) .. (319.25,110.75) -- (194.25,110.75) .. controls (188.86,110.75) and (184.5,106.39) .. (184.5,101) -- cycle ;
\draw    (40.1,94.73) -- (40,18) ;
\draw [shift={(40,15)}, rotate = 89.93] [fill={rgb, 255:red, 0; green, 0; blue, 0 }  ][line width=0.08]  [draw opacity=0] (10.72,-5.15) -- (0,0) -- (10.72,5.15) -- (7.12,0) -- cycle    ;
\draw    (35.6,89.98) -- (167,90) ;
\draw [shift={(170,90)}, rotate = 180.01] [fill={rgb, 255:red, 0; green, 0; blue, 0 }  ][line width=0.08]  [draw opacity=0] (10.72,-5.15) -- (0,0) -- (10.72,5.15) -- (7.12,0) -- cycle    ;
\draw [color={rgb, 255:red, 74; green, 144; blue, 226 }  ,draw opacity=1 ][line width=1.5]    (40,90) .. controls (99.5,58.25) and (102.08,52.71) .. (148.75,41.38) ;
\draw [color={rgb, 255:red, 0; green, 0; blue, 0 }  ,draw opacity=0.5 ] [dash pattern={on 0.84pt off 2.51pt}]  (40,40) -- (150,40) ;
\draw [color={rgb, 255:red, 208; green, 2; blue, 27 }  ,draw opacity=1 ] [dash pattern={on 4.5pt off 4.5pt}]  (40,80) -- (150.07,80.2) ;
\draw    (150.07,86.87) -- (150,92.53) ;
\draw [color={rgb, 255:red, 65; green, 117; blue, 5 }  ,draw opacity=1 ]   (118.92,87.12) -- (118.92,92.78) ;
\draw [color={rgb, 255:red, 65; green, 117; blue, 5 }  ,draw opacity=1 ][line width=1.5]    (40,90) .. controls (127,89.25) and (103.33,81.58) .. (150,70.25) ;
\draw [color={rgb, 255:red, 74; green, 144; blue, 226 }  ,draw opacity=1 ]   (58.42,87.12) -- (58.42,92.78) ;
\draw [color={rgb, 255:red, 74; green, 74; blue, 74 }  ,draw opacity=0.6 ][line width=1.5]  [dash pattern={on 3.75pt off 3pt on 7.5pt off 1.5pt}]  (40,90) .. controls (117.8,82.7) and (70.8,75.7) .. (150.6,57.55) ;
\draw [color={rgb, 255:red, 74; green, 144; blue, 226 }  ,draw opacity=1 ][line width=1.5]    (189,15.75) -- (209,5.25) ;
\draw [color={rgb, 255:red, 74; green, 74; blue, 74 }  ,draw opacity=0.5 ][line width=1.5]  [dash pattern={on 3.75pt off 3pt on 7.5pt off 1.5pt}]  (189,51.25) -- (209,40.75) ;
\draw [color={rgb, 255:red, 65; green, 117; blue, 5 }  ,draw opacity=1 ][line width=1.5]    (189.5,91.25) -- (209.5,80.75) ;

\draw (29,33.5) node [anchor=north west][inner sep=0.75pt]  [font=\footnotesize] [align=left] {1};
\draw (146.33,94.9) node [anchor=north west][inner sep=0.75pt]  [font=\footnotesize]  {$\lambda _{\max}$};
\draw (45.5,6) node [anchor=north west][inner sep=0.75pt]  [font=\footnotesize]  {$\mathbb{P}[\mathrm{coll}_{i}(\mathsf{C}( E,\lambda ))]$};
\draw (29.33,92) node [anchor=north west][inner sep=0.75pt]  [font=\footnotesize] [align=left] {0};
\draw (30.5,69.33) node [anchor=north west][inner sep=0.75pt]  [font=\footnotesize] [align=left] {$\displaystyle \textcolor[rgb]{0.82,0.01,0.11}{\epsilon }$};
\draw (110.5,94.9) node [anchor=north west][inner sep=0.75pt]  [color={rgb, 255:red, 65; green, 117; blue, 5 }  ,opacity=1 ]  {$\overline{\lambda }^{*}$};
\draw (158.33,71.73) node [anchor=north west][inner sep=0.75pt]  [font=\footnotesize]  {$\lambda $};
\draw (49.5,92.9) node [anchor=north west][inner sep=0.75pt]  [color={rgb, 255:red, 74; green, 144; blue, 226 }  ,opacity=1 ]  {$\underline{\lambda }^{*}$};
\draw (211.67,2.83) node [anchor=north west][inner sep=0.75pt]  [font=\footnotesize] [align=left] {Safety margin curve\\with non-reactive AV};
\draw (211.92,79) node [anchor=north west][inner sep=0.75pt]  [font=\footnotesize] [align=left] {Safety margin curve\\with avoidance policy};
\draw (212,40) node [anchor=north west][inner sep=0.75pt]  [font=\footnotesize] [align=left] {Unknown real safety\\margin curve};

\end{tikzpicture}